\documentclass[journal]{IEEEtran}
\newcommand{\ignore}[1]{}

\usepackage[utf8]{inputenc}
\usepackage[T1]{fontenc}
\usepackage{amsmath,epsfig}
\usepackage{url}
\usepackage{graphicx}
\usepackage{bm}
\usepackage{amssymb}
\usepackage{cite}
\usepackage{array}
\usepackage{color}
\usepackage{booktabs}
\usepackage{multirow}

\usepackage{times}
\usepackage{epsfig}
\usepackage{graphicx,psfrag,epsfig}
\usepackage{amsmath}
\usepackage{amssymb}

\usepackage{multirow}
\usepackage{array}
\usepackage{caption}
\usepackage{makecell}

\begin{document}

%%%%%%%%% TITLE
\title{Spatial Semantic Recurrent Mining for Referring Image Segmentation}

\author{Jiaxing Yang, Lihe Zhang, Jiayu Sun, Huchuan Lu\\
\IEEEcompsocitemizethanks{
\IEEEcompsocthanksitem This work was supported by the National Key R\&D Program of China \#2018AAA0102003, the National Natural Science Foundation of China  \#62276046, and the Liaoning Natural Science Foundation  \#2021-KF-12-10.
\IEEEcompsocthanksitem  J. Yang, L. Zhang, J. Sun, and H. Lu are
  with School of Information and Communication Engineering, Dalian
  University of Technology, Dalian, China.
.
}
}
\markboth{}{}

\maketitle
%%%%%%%%% ABSTRACT
\begin{abstract}
Referring Image Segmentation (RIS) consistently requires language and appearance semantics to more understand each other. The need becomes acute especially under hard situations. To achieve, existing works tend to resort to various trans-representing  mechanisms to directly feed forward language semantic along main RGB branch, which however will result in referent distribution weakly-mined in space and non-referent semantic contaminated along channel. In this paper, we propose Spatial Semantic Recurrent Mining (S\textsuperscript{2}RM) to achieve high-quality cross-modality fusion. It follows a working strategy of trilogy:  distributing language feature, spatial semantic recurrent coparsing, and parsed-semantic balancing.  During fusion, S\textsuperscript{2}RM will first generate a constraint-weak yet distribution-aware language feature, then bundle features of each row and column from rotated features of one modality context to recurrently correlate relevant semantic contained in feature from other modality context, and finally resort to self-distilled weights to weigh on the contributions of different parsed semantics. Via coparsing, S\textsuperscript{2}RM transports information from the near and remote slice layers of generator context to the current slice layer of parsed context, capable of better modeling global relationship bidirectional and structured. Besides, we also propose a Cross-scale Abstract Semantic Guided Decoder (CASG) to emphasize the foreground of the referent, finally integrating different grained features at a comparatively low cost. Extensive experimental results on four current challenging datasets show that our proposed method performs favorably against other state-of-the-art algorithms.  
\end{abstract}


\IEEEpeerreviewmaketitle
%%%%%%%%% BODY TEXT
\section{Introduction}
\label{sec:intro}
Along with requirement surging on human-robot interaction, cross-modality fusion recently has aroused wide attention among the vision community. The involved tasks include video-text retrieval~\cite{chen2020fine}, image captioning~\cite{lu2017knowing, xu2015show}, visual question answering~\cite{lu2016hierarchical, nie2023temporal}, and language-guided object segmentation/grounding~\cite{hu2016segmentation, li2018referring}. In this paper, we focus our attention on referring image segmentation task (RIS). Different from instance segmentation method grouping different pixels into distinct instances, RIS requires the networks to have a more fine-grained understanding on image content, and to segment out referred region according to given expression, one or more. Usually, the region is either stuff or object. The expression describes target's action, category, color, position, and \emph{etc}. As a relatively novel topic, RIS faces lots of problems to solve. One is that the task requires the proposed methods to process multiple types of information appearing in two modalities. Another is that the feature embeddings from the vision and language modalities should approach each other as close as possible, for the networks to generate expressive cross-modality feature.

Benefiting from the development of deep learning based techniques, early methods achieve the task by first resorting to CNN~\cite{cnn}, LSTM~\cite{lstm}, GRU~\cite{gru} based techniques to extract related features, and then fusing the features in a coarse way. As the first one posting RIS problem, the authors in ~\cite{hu2016segmentation} use VGG~\cite{vgg} and LSTM~\cite{lstm} to extract the vision and language features, respectively, then concatenate the features together, and finally decode the referent appearance via a fully convolutional pixel-wise classification structure. The Recurrent Refinement Network (RRN)~\cite{li2018referring} applies a ConvLSTM decoder to gradually recover details of the referred region, which is able to process multi-scale features extracted using ResNet~\cite{he2016deep}. Different from their sentence-level reasoning strategy, RMI~\cite{liu2017recurrent} achieves cross-modality fusion by explicitly modeling word-to-image interaction process in a recurrent manner. The ignorance of above mentioned methods towards contributions of different words finally drags down their performance, especially for complex scenarios.

To resolve, many following works~\cite{shi2018key, yelinwei2019cross, hui2020linguistic, feng2021encoder, vlt, hu2020bi123, yu2018mattnet, luo2020multi, lts, Kim2022ReSTRCR, mdetr, Liu2022InstanceSpecificFP} have designed new fusion mechanisms to achieve more impressive referring performance. In MAttNet~\cite{yu2018mattnet}, Faster RCNN~\cite{fasterrcnn} is first employed to propose regions of interest, and then the most relevant referent mask is selected by a designed Modular Attention Network. The IFPNet~\cite{Liu2022InstanceSpecificFP} designs a unified architecture to first predict the existence of the referent and then accordingly predicts its corresponding mask, quite inspired by SOLO~\cite{solo1} and its follow-ups. Nevertheless, their latent assumption treating referent as instance is problematic. In LTS~\cite{lts}, a localization module is proposed to primarily obtain visual prior according to a coarse cross-modality feature map. Then the attention cue is used to instruct the final segmentation process of referent. However, the final results by LTS~\cite{lts} may be dragged down by poor cues. The BRINet~\cite{hu2020bi123} proposes to use bidirectional cross-modality attention module to model the relationship of cross modality features. The CEFNet~\cite{feng2021encoder} proposes to use a co-attention embedding module to update multi-modal features by instilling human-specified heuristic experience. In general, these methods~\cite{shi2018key, yelinwei2019cross, hui2020linguistic, feng2021encoder, vlt, hu2020bi123, yu2018mattnet, luo2020multi, lts, Kim2022ReSTRCR, mdetr, Liu2022InstanceSpecificFP} heavily rely on CNN-based techniques, therefore constrained by the innate drawback of 2D convolution manipulation. Usually, static kernel parameter owns weak representative ability. 

Recent pioneering works~\cite{Kim2022ReSTRCR, yang2022lavt,yang2023semantics} introduce transformer-based techniques into referent segmentation. Dynamic content-aware weights not only can more effectively capture the complexity of image content from a large-scale dataset but also model inter-modality relationship. Inspired by ~\cite{attention1, vit, swin}, ReSTR~\cite{Kim2022ReSTRCR} resorts to cross-attention mechanism to fuse cross-modality features in patch level, and the product is purified by a language-guided weighting map before sent to a pixel-classified decoder. However, patch-based classification in the convolution-free ReSTR may cause the network unable to fully utilize the multiscale semantics in vision feature extracting process. Quite inspired by CEFNet~\cite{feng2021encoder}, LAVT~\cite{yang2022lavt} successfully introduces early fusion strategy into feature extracting stage of Swin transformer. Over LAVT, SADLR~\cite{yang2023semantics} functions by introducing cascaded decoding paradigm.  

Overall, both first-tile-then-concatenation and vanilla seq2seq aligning~\cite{nonlocal} based solutions are plagued by inductive bias and semantic contamination headaches. The former means that the network understanding the abstract linguistic semantic more relies on local-semantic strong vision tokens. The reconstructed language token tends to be independent to each other, lacking of structured constraint and thus weak in discerning environment and distribution information within/around referent. The latter signifies that the non-referred region is also expressed by the language semantics, more or less. Their following accommodation towards the pure vision feature embedding and its corresponding retrofitted language embeddings directly uses bluff linear recombination. Therefore, semantic contamination in channel is aroused. Admittedly, early fusion mechanism lets the above mentioned problems alleviated to some extent, as the 'Deep' design in network architecture is important (prove in ResNet~\cite{he2016deep}). However, it sacrifices a lots. When several referents need to be delineated out, the extracted features can not be reused by other expressions. Moreover, its interruptions on paranoid transformer-based backbones also afflict the prediction process.

To resolve, we achieve fine-grained cross-modality fusion by proposing Spatial Semantic Recurrent Mining (S\textsuperscript{2}RM), and integrate multiscale semantics by engineering Cross-scale Abstract Semantic Guided decoder (CASG), which operates at a comparatively low cost. To reduce exotic interrupts and enhance feature reusibility, we install S\textsuperscript{2}RM on top of feature extractors. The S\textsuperscript{2}RM consists of distributing language feature, spatial semantic recurrent coparsing, and parsed-semantic balancing steps. During fusion, S\textsuperscript{2}RM will primarily produce a distribution-aware language feature, then bundle each row and column of rotated features from one modality to recurrently parse relevant semantic contained in feature from the other modality, and finally utilize self-distilled weights to balance the contributions of parsed semantics. In the second step, information from the near and remote slice layers of generator context is transported into the current slice layer of parsed context, finally modelling global relationship in a bidirectional (Language$\&$Vision) and structured (Column$\&$Row) way. The CASG resorts to language feature and cross-scale features from the previous decoding stages to achieve adaptive vision feature purification. Within, spatial and channel attention guidance are generated to enlighten the uncivilized vision feature to more amplify the described content. Contributions are: 

\begin{itemize}
	\item We design Spatial Semantic Recurrent Mining (S\textsuperscript{2}RM) for referring image segmentation to achieve effective interaction of vision/language features. A trilogy strategy is employed to promote their mutual understanding. 

    \item The S\textsuperscript{2}RM transports information from the near and remote slice layers of one context to the corresponding slice layer of current context, modeling global relationship of different embeddings bidirectional and structured.
	
    \item We propose Cross-scale Abstract Semantic Guided Decoder (CASG) to gradually integrate multiscale features. In the process, it computes spatial and channel attention using purified high abstract features and language embedding to provide refined referent mask.
	
	\item We install S\textsuperscript{2}RM on top of the feature extractor and in addition resort to CASG in the decoder to supplement the details of referent, achieving favorable results against other state of the arts on four current challenging datasets, RefCOCO, RefCOCO+, RefCOCOg, and ReferIt. 
\end{itemize}

\section{Related Works}
\label{sec:relatedWorks}
\textbf{Semantic Segmentation.} Semantic segmentation~\cite{long2015fully,chen2017deeplab, chen2017rethinking, unet, badrinarayanan2017segnet} will categorize each pixel of image into different pre-specified semantic categories. The FCN~\cite{long2015fully} is the primary one proving feasibility of per-pixel classification in general scenario segmentation. Contemporary UNet~\cite{unet}, as well as ~\cite{deconvnet, ding2018context}, in medical image segmentation develops skip connection across encoder and decoder, and gradually integrates the features of primary stages to refine segmentation process. Deeplab serious~\cite{chen2017deeplab, chen2017rethinking} enhance global information gleaning ability of CNN by introducing atrous rate into 2D convolution, and employ the design in a pyramidal manner. The FuseNet~\cite{hazirbas2016fusenet} integrates depth information in the encoder to conduct semantic segmentation. The authors in \cite{chen2020bi} design SAGate module to fuse cross-modality features in a bidirectional way. The MS-APS-Net ~\cite{zhao2021multi} proposes to orchestrate optical, depth, RGB, image information to segment out dynamic foreground. Recently, nonlocal based ViT~\cite{vit} is employed to accommodate the variety of image content in semantic segmentation. After projecting full images into patch embeddings, information is communicated across the patches via a stack of transformer blocks. Its follow-ups Swin transformer~\cite{swin} resorts to shifting window mechanism to dynamically interact information in a multi-scale manner. The PVT~\cite{pvt} adapts seq2seq transformer by heralding spatial reduction on key and value entities to save calculation overhead of affinity matrices.

\textbf{Instance Segmentation.} Compared with semantic segmentation, instance segmentation methods~\cite{he2017mask, liao2020real, solo1} in general own a more fine-grained understanding towards image contents. They will segment out the appearance of each instance in the image. One group of them first boxes the position of instance, and then resorts to FCN~\cite{long2015fully} structure to segment out instance appearance, following a two-step paradigm. Among the methods, mask RCNN~\cite{he2017mask} is the most classical solution. It first localizes the instance via Faster RCNN~\cite{fasterrcnn} and then resorts to multiple convolution layers to conduct binary segmentation. Many follow-ups of Mask RCNN, e.g., YOLO~\cite{liao2020real} and DETR~\cite{detr}, spring up advancing the task by instilling their own observations. Another group of them~\cite{solo1, wang2020solov2, detr, xie2020polarmask} abandon the localization step and directly predict the instance mask according to instance existence. The mechanisms of feature pyramid~\cite{lin2017feature}, feature aggregation~\cite{liu2018path}, and cascaded optimization~\cite{chen2019hybrid} are also reviewed.

\textbf{Language Encoding.} The seq2seq transformer~\cite{transformer} is the milestone one starting a new era, which at first is proposed to solve the long-range information loss, gradient missing, gradient boom problems existing in LSTM~\cite{lstm} and GRU~\cite{gru}. Then Transformer gradually occupies a variety of language and vision tasks given its powerful modeling ability on inner/inter sequences, in which Non-local attention manipulation~\cite{nonlocal} plays a pivotal role. Encoder and decoder of Transformer is bifurcated into Bert~\cite{bert} and GPT~\cite{gptv1}, respectively. The GPT and its variants have been unifyingly named as Large Model, which shows powerful emerging ability and has been regarded as the most potential candidate in achieving artificial intelligence. The Clip~\cite{clip} designs a contrasting loss to coordinate the correspondence of image-text pair in training, finally obtaining more powerful encoding architecture unifying two different modalities. Note that the model is the primary one pretrained on large-scale cross-modality datasets, which consists of four million image-text pairs. 

\textbf{Referring Image Segmentation.} Different from instance segmentation, referring image segmentation methods will segment out the referent appearance described by an expression, a task that is able to achieve more nimble segmentation in terms of target. Early of them like~\cite{hu2016segmentation, liu2017recurrent, li2018referring, dmn} adopt a first-tile-then-concatenate strategy, which is hard trans-parsed and is able to handle some easy scenarios. In \cite{hu2016segmentation}, the task of RIS is posed. The authors resort to VGG~\cite{vgg} and LSTM~\cite{lstm} to extract vision and language features, respectively. The sentence semantic of language is tiled and attached to vision feature for FCN~\cite{long2015fully}-like segmentation head. Different from \cite{hu2016segmentation}, RMI~\cite{liu2017recurrent} fuses each word embedding into vision feature in a recurrent way. The KWAN~\cite{shi2018key} uses query attention module to extract the most relevant linguistic semantic for each image region, and meanwhile employs key-word-aware visual context module to model key-word-attended visual context semantic. An MLP segmentation head feeding on key-word-aware visual context features, key-word-weight query features is used to predict the final mask. The LSCM~\cite{hui2020linguistic} updates constructed graph to remove the disturbing cues before final segmentation process. In BRINet~\cite{hu2020bi123}, a bidirectional cross-modality attention module is used to model the relationship of cross-modality features, and a gated bidirectional fusion module is applied to achieve more impressive performance. In CEFNet~\cite{feng2021encoder}, a co-attention fusion module is used to update cross-modal features, which is employed in the early vision feature extraction process to achieve deep fusion. Still in CEFNet~\cite{feng2021encoder}, a boundary enhancement module is developed to achieve finer segmentation.  In VLT~\cite{vlt}, appearance cues are incorporated into query generation process to enhance holistic understanding, which then is used to generate weights vector to amplify the foreground semantic before decoding.

In MAttNet~\cite{yu2018mattnet}, a Modular Attention Network is proposed to select referent mask from Faster RCNN~\cite{fasterrcnn}. The MCN~\cite{luo2020multi} proposes consistency energy maximization and adaptive soft non-located suppression to conduct multi-task learning so as to assist referent segmentation. The LTS~\cite{lts} first uses a localization module to directly obtain the visual prior based on a coarse cross-modality feature map, and then draws on the prior to segment out the referred region. The ISPNet~\cite{Liu2022InstanceSpecificFP} proposes a two-branch 
strategy to in one branch determine existence of referred instance and to in the other branch select the corresponding mask, quite motivated by SOLO~\cite{solo1}. In PolyFormer~\cite{liu2023polyformer}, the segmentation of referent are parsed as prediction of sequential points. In LAVT~\cite{yang2022lavt}, a gating mechanism is designed to infuse the language into Swin transformer. Over LAVT, SADLR~\cite{yang2023semantics} gradually introduces more-refined localization cues to decode. In CRIS~\cite{wang2022cris}, Clip-pretrained model is employed to achieve supervised segmentation driven by a designed contrastive loss. In Zero-Shot RIS~\cite{yu2023zero}, vision and language encoder in Clip is adapted to achieve unsupervised RIS.
In MCRES~\cite{xu2023meta}, a meta optimization scheme is engineered to handle novel compositions of learned concepts.
In TRIS~\cite{liu2023referring}, language information is used to achieve weakly RIS, and a calibration strategy is applied to generate high-quality response maps.

\begin{figure*}[htbp]
	\centering
	\includegraphics[scale=0.68]{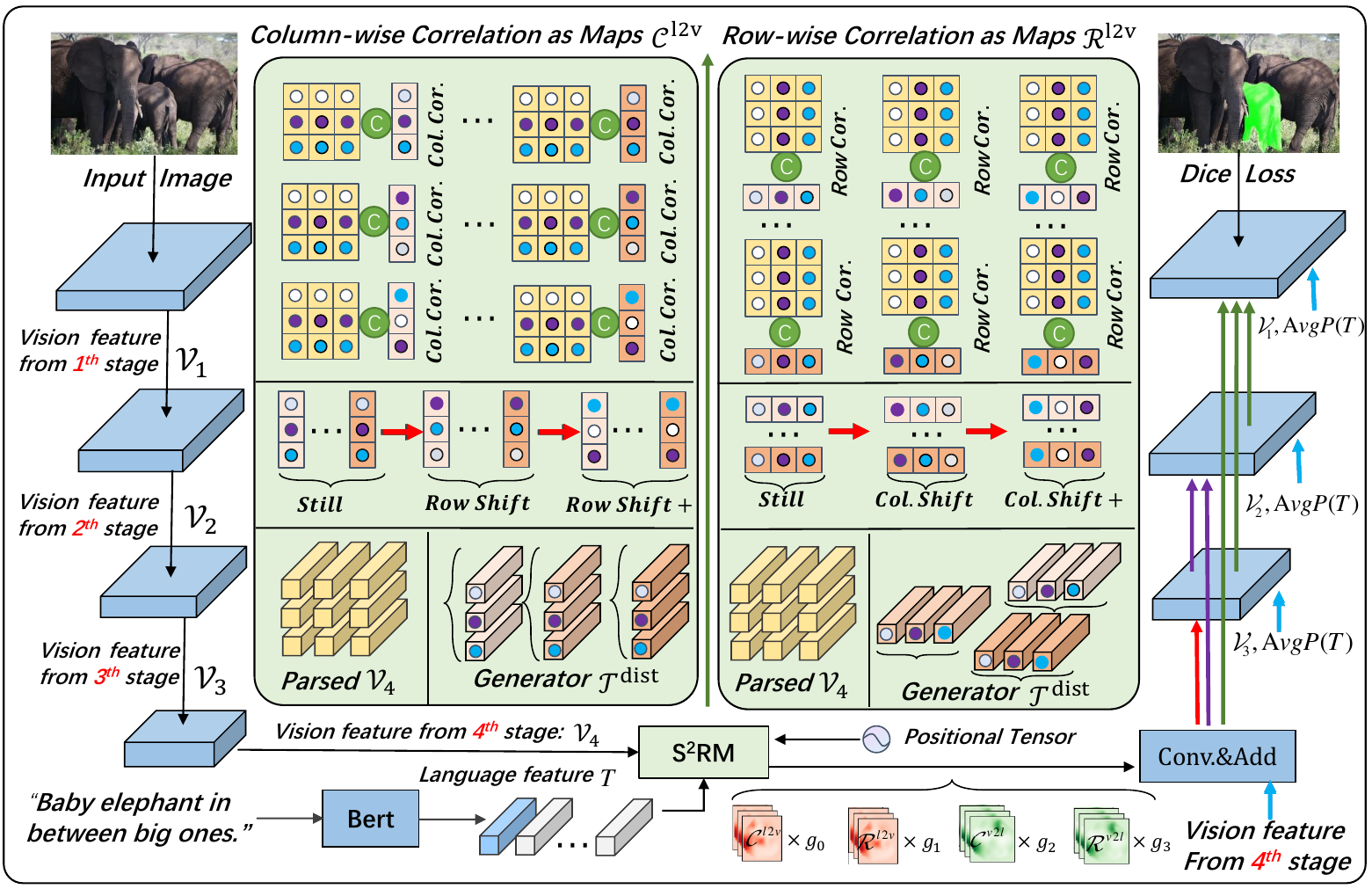}
	\caption{Architecture overview. On top of Swin and Bert, the proposed S\textsuperscript{2}RM is installed to mine global distribution information in a bidirectional and structured way. Middle green parts (second step of S\textsuperscript{2}RM) visualize how to use generator context $\mathcal{T}^{\rm dist}$ to generate content-adaptive slices to correlate parsed context $\mathcal{V}_{4}$ as maps in column-wise and row-wise. }
	\label{Architecture}
\end{figure*}
%------------------------------------------------------------------------
\section{Methodology}
\label{sec:methods}
This section will in order detail the overall architecture of our network, the proposed Spatial Semantic Recurrent Mining (S\textsuperscript{2}RM), as well as the designed Cross-scale Abstract Semantic Guided decoder (CASG). 

\subsection{Overall Architecture}
Previous vision semantic extraction mainly relies on CNN, for example VGG~\cite{vgg}, ResNet~\cite{he2016deep}, DarkNet~\cite{liao2020real}, and ResNest~\cite{resnest}. However, there are two problems facing them. The first is that CNN based extractors have limited local and global information gleaning ability, which is caused by their consideration over controlling parameter number. The second is that static parameter is plagued by their weak representative ability. The stacked 2D convolutional manipulation can not be adaptive according to the various input. In this work, we use Swin transformer serious backbone to extract vision semantic, and 12-layer Bert base to process relevant expression, respectively.  We visualize the overall structure of our network in Fig. \ref{Architecture}. It can be seen that the developed S\textsuperscript{2}RM is only installed on top of feature extractors. Usually, one image is bundled with multiple expressions. Different from early fusion in CEFNet~\cite{feng2021encoder}, LAVT~\cite{yang2022lavt} and SADLR~\cite{yang2023semantics}, our adopted fusion strategy avoids semantic from one expression contaminating the downstreaming feature extracting process, and the extracted feature in this way can be reused by other expressions related to the image content.  Features in the early stages also should be integrated into the segmentation process. Therefore, we propose a lightweight decoder without resorting to nonlocal~\cite{nonlocal} variants. In the process, one cross-modality feature, three pure vision features, sentence-level language embedding are together fed to decoder CASG. 

For mathematical tractability, calligraphic uppercase letter, italic uppercase letter, nonitalic uppercase letter, and italic lowercase letter are used to represent tensor, matrix, constant number, and variable, respectively. The superscript of its corresponding main body symbol specifies its semantic identity. Let $\cal{V}$ of size $\rm H\times \rm W\times \rm C$ represent image, in which $\rm H$, $\rm W$, and $\rm C$ are the height, width, and channel dimension, respectively. And $T$ of size $\rm N\times \rm C^{lang}$  is accepted to represent word embedding after Bert-base encoding, where $\rm N$ and $\rm C^{lang}$ represent language length and embedding dimension, respectively. In the $i$th stage, vision feature is denoted as $\mathcal{V}_{i}$ with size of $\rm{H}/2^{(\it{i}+\rm{1})} \times \rm{W}/2^{(\it{i}+\rm{1})} \times \rm{C}_{\it{i}}^{vis}$, where $i=1,...,4$. 

\begin{figure}[tbp]
	\centering
	\includegraphics[scale=0.65]{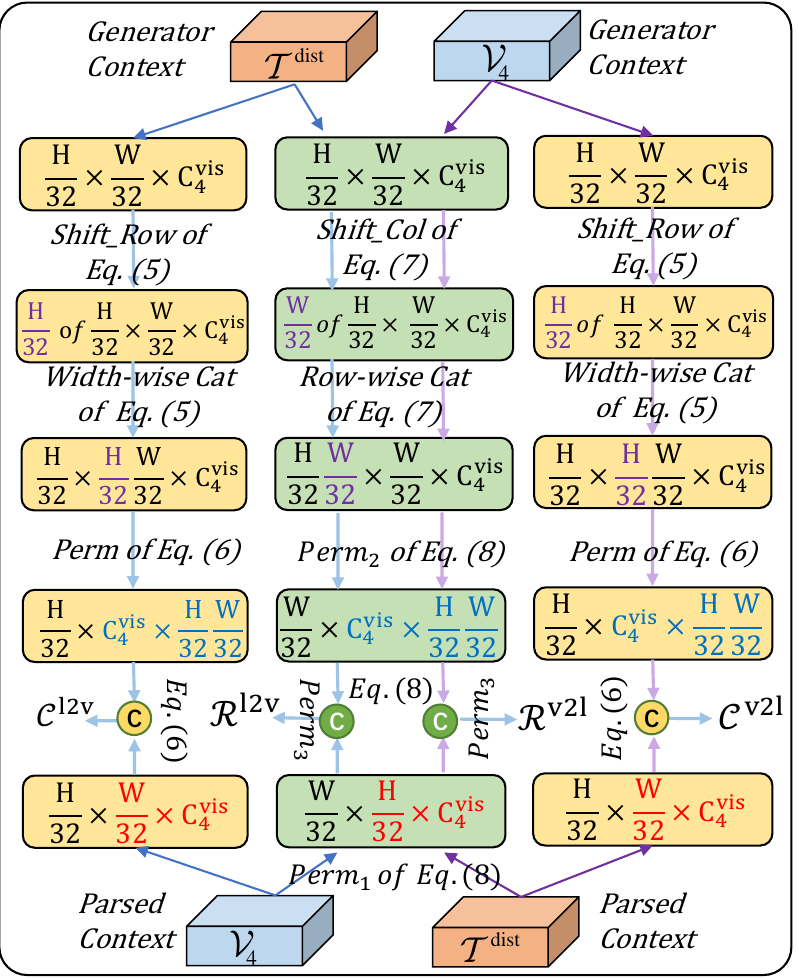}
	\caption{Transformations in Spatial Semantic Recurrent Coparsing of S\textsuperscript{2}RM, where slice layers from one modality parse semantics from other modality as four group maps.}
	\label{VSP}
\end{figure}
\subsection{Spatial Semantic Recurrent Mining}

The usage of S\textsuperscript{2}RM guarantees extraction stabilized, overhead controlled, and feature representation ability strong. The S\textsuperscript{2}RM goes through three steps in order to generate fine-grained cross-modality fusion product: Distributing Language Feature, Spatial Semantic Recurrent Coparsing, and Parsed Semantics Balancing. The first step aims to generate a local-semantic-strong distribution-aware language feature. Instead of direct feeding forward summation of vision and language embeddings, the second step mines the referent distribution information in a bidirectional and structured way. This step processes pure vision feature $\mathcal{V}_4$ and the language feature to alleviate the aforementioned problems. \textbf{During the process, content-adaptive slices structured along either row or column from one modality feature, will be generated to correlate the feature from other modality according to column-wise and row-wise Hadamand product, respectively}. Hereafter, the feature generating kernels, and the feature to be parsed are called as generator context and parsed context, respectively. In the correlation process, row and column slices of generator context are cyclically shifted to augment global representational ability along horizontal and vertical directions, without the need to introduce blank placeholder. The third step weighs on contribution significance of different parsed semantics considering referent sprawling.

\subsubsection{Distributing Language Feature}
In this step, a coarse language-aware vision feature primarily is generated as:
\begin{equation} \label{q}
	\begin{aligned}
		& \mathcal{Q} = Conv2D\_BN\_ReLU(Cat(\mathcal{V}_4,\mathcal{T}_0,\mathcal{P})),
	\end{aligned} 
\end{equation}
where $\mathcal{V}_4$ is the pure vision feature extracted on top of Swin transformer; The $\mathcal{T}_0$ is generated resorting to the $Tile$ manipulation to make vector $AvgP(T)$ keep aligned to $\mathcal{V}_4$ in spatial dimension, where $AvgP()$ represents average pooling; The positional tensor $\mathcal{P}$ of size $\rm{H}/32\times \rm{W}/32 \times 8$ is generated following previous works~\cite{hu2016segmentation, hu2020bi123, feng2021encoder}. The manipulation of $Cat$ denotes to concatenate the tensors along their channel dimension. The $Conv2D\_BN\_ReLU$ represents to use a stack of $1\times 1$ 2D Convolution, BatchNorm, and ReLU to compress the channel dimension of concatenated product to $\rm{C}^{\rm{vis}}_{4}$. The query and key entities can be generated as:
\begin{equation} \label{kv}
	\begin{aligned}
        & K = Perm(Linear\_LN\_ReLU(T)),\hspace{3pt} \\
		& V =Linear\_LN\_ReLU(T),\hspace{3pt}
	\end{aligned} 
\end{equation}
where the $Linear\_LN\_ReLU$ represents to use a stack of 1D convolution, LayerNorm, and ReLU to adjust the channel dimension of language embedding to $\rm{C}^{\rm{vis}}_{4}$. The manipulation of $Perm$ represents to exchange the first and second indices of the manipulated tensor. Therefore the query and key entities can be utilized to generate the following affinity matrix:
\begin{equation} \label{att}
	\begin{aligned}
		&A = Softmax((Flatten(\mathcal{Q})\otimes K)/\sqrt{\rm{C^{vis}_4}},
	\end{aligned} 
\end{equation}
in which the manipulation of $Flatten$ represents to unroll the feature in a first-row-then-column manner, the symbol of $\otimes$ represents matrix multiplication, and the manipulation $Softmax()$ represents exponential normalization along the second dimension. The affinity matrix $A$ is of size $\rm{(H/32)(W/32)}\times \rm{N}$. By gleaning the information over the value entity $V$, we can use the matrix to generate the following distribution aware language tensor:
\begin{equation} \label{cor}
	\begin{aligned}
		\mathcal{T}^{\rm{dist}}= Reshape(A\otimes V),
	\end{aligned} 
\end{equation}
in which the feature $\mathcal{T}^{\rm dist}$ is of size $\rm{H/32\times W/32\times \rm{C}^{vis}_4}$. The manipulation $Reshape$ represents to recover the unrolled feature to keep aligned to the original vision feature.

\subsubsection{Spatial Semantic Recurrent Coparsing}
In this step, $\mathcal{V}_4$ and $\mathcal{T}^{\rm dist}$ are utilized to generate four groups of cross-modality feature maps as column-wise language-attended vision feature $\mathcal{C}^{\rm{l2v}}$, row-wise language-attended vision feature $\mathcal{R}^{\rm{l2v}}$, column-wise vision-attended language feature $\mathcal{C}^{\rm{v2l}}$, and row-wise vision-attended language feature $\mathcal{R}^{\rm{v2l}}$. We refer the readers to attend to green region of Fig. \ref{Architecture} to follow the upcoming mathematical generation of $\mathcal{C}^{\rm{l2v}}$ and $\mathcal{R}^{\rm{l2v}}$ from Eqs. \eqref{l2v_row1} to \eqref{l2v_col2}. The shape transformation of the involved high-order tensors is referred to Fig. \ref{VSP}. Note that the features $\mathcal{C}^{\rm{v2l}}$ and $\mathcal{R}^{\rm{v2l}}$ can be obtained by exchanging position of $\mathcal{V}_4$ and $\mathcal{T}^{\rm dist}$ in Eqs. \eqref{l2v_row1}\eqref{l2v_row2} and \eqref{l2v_col1}\eqref{l2v_col2}, namely $func(a, b)->func(b,a)$.

To be specific, generating $\mathcal{R}^{\rm{l2v}}$ requires to process the generator context $\mathcal{T}^{\rm{dist}}$ as follows: 
\begin{equation} \label{l2v_row1}
	\begin{aligned}
		\mathcal{T}^{\rm{row}} = Cat(&\mathcal{T}^{\rm{dist}}, Shift\_Row(\mathcal{T}^{\rm{dist}}, 1), ..., \\
                          &Shift\_Row(\mathcal{T}^{\rm{dist}}, h), ..., \\
                          &Shift\_Row(\mathcal{T}^{\rm{dist}}, \rm{H/32}-1)),
	\end{aligned} 
\end{equation}
in which $h=1,...,\rm{H/32}-1$; the manipulation $Cat(\cdot)$ represents to concatenate the tensors along the specified dimension, namely along the width; The manipulation of $Shift\_Row$ represents to cyclically shift the row slices of feature given the moving step. For instance, the $i$th row slice of $\mathcal{T}^{\rm{dist}}$ before transformation will correspond to the $Mod(i+h, \rm{H/32})$th row slice of $Shift\_Row(\mathcal{T}^{\rm{dist}}, h)$ after transformation, where $Mod(a,b)$ represents Modulo Operation of $a$ with respect to $b$. The column-slice-set kernel tensor $\mathcal{T}^{\rm{row}}$ is of size $\rm{(H/32)}\times \rm{(H/32)}\rm{(W/32)} \times \rm{C_4^{vis}}$. 

Therefore, $\mathcal{C}^{\rm{l2v}}$ of size $\rm{H/32\times W/32 \times (H/32)(W/32)}$ can be generated by the following correlation manner:
\begin{equation} \label{l2v_row2}
	\begin{aligned}
		\mathcal{C}^{\rm{l2v}} = \mathcal{V}_4\otimes Perm(\mathcal{T}^{\rm{row}}),
	\end{aligned} 
\end{equation}
in which the manipulation $Perm$ changes the indices of $\mathcal{T}^{\rm{row}}$ from $0,1,2$ to $0,2,1$ to accommodate the fusion process. The symbol of $\otimes$ corresponds to the correlation manipulation in Fig. \ref{Architecture} and Fig. \ref{VSP}, by which information sinked in the channel dimension of parsed context $\mathcal{V}_4$ is translated as maps of $(\rm H/32)(\rm W/32)$.

Generating $\mathcal{R}^{\rm{l2v}}$ requires to process $\mathcal{T}^{\rm{dist}}$ as:
\begin{equation} \label{l2v_col1}
	\begin{aligned}
		\mathcal{T}^{\rm{col}} = Cat(&\mathcal{T}^{\rm{dist}}, Shift\_Col(\mathcal{T}^{\rm{dist}}, 1), ..., \\
                            &Shift\_Col(\mathcal{T}^{\rm{dist}}, w), ..., \\
                            &Shift\_Col(\mathcal{T}^{\rm{dist}}, \rm{W/32}-1)),
	\end{aligned} 
\end{equation}
in which $w=1,...,\rm{W/32}-1$; the manipulation $Cat(\cdot)$ represents to concatenate the tensors along the height dimension; The manipulation of $Shift\_Col$ represents to cyclically shift the column slices of feature according to the specified moving step. The $i$th column slice of $\mathcal{T}^{\rm{dist}}$ before transformation will correspond to the $Mod(j+w, \rm{W/32})$th column slice of $Shift\_Col(\mathcal{T}^{\rm{dist}}, w)$ after transformation. The row-slice-set kernel tensor $\mathcal{T}^{\rm{col}}$ is of size $\rm{(H/32)(W/32)}\times \rm{W/32} \times \rm{C_4^{vis}}$. Therefore, $\mathcal{R}^{\rm{l2v}}$ of size $\rm{H/32\times W/32 \times (H/32)(W/32)}$ can be generated by the following correlation:
\begin{equation} \label{l2v_col2}
	\begin{aligned}
		\mathcal{R}^{\rm{l2v}} = Perm_3(Perm_1(\mathcal{V}_4)\otimes Perm_2(\mathcal{T}^{\rm{col}})),
	\end{aligned} 
\end{equation}
in which the manipulation $Perm_1$ changes the indices of $\mathcal{V}_4$ from $0,1,2$ to $1,2,0$; the manipulation $Perm_2$ changes the indices of $\mathcal{V}_4$ from $0,1,2$ to $1,0,2$; $Perm_3$ represents to change the indices of $\mathcal{V}_4$ from $0,1,2$ to $1,0,2$. 

\subsubsection{Parsed Semantics Balancing}
Before instilling the scanned maps into the main branch, we adopt a rebalancing method to more identify effective information facilitating the final segmentation process. We first use the four groups of features to calculate four different weights. The process can be denoted as follows:
\begin{equation} \label{stage3}
	\begin{aligned}
		&\mathcal{G} = Cat(\mathcal{C}^{\rm{l2v}}, \mathcal{R}^{\rm{l2v}}, \mathcal{C}^{\rm{v2l}}, \mathcal{R}^{\rm{v2l}})\\
        & G = Sig(AvgP(Conv(\mathcal{G}))),
	\end{aligned} 
\end{equation}
in which the manipulation $Cat$ represents to concatenate the tensors along their third dimension; The $Conv$ represents to compress the channel dimension to four planes; the $AvgP()$ represents to conduct average pooling over the spatial dimension; The $Sig$ represents to apply sigmoid manipulation. The $G$ is a vector with size $4$, which can be expressed as $[g_0, g_1,g_2,g_3]$. Then the weights can be used as follows:
\begin{equation} \label{skip3}
	\begin{aligned}
		\mathcal{B} = Cat([\mathcal{C}^{\rm{l2v}}*g_0, \mathcal{R}^{\rm{l2v}}*g_1, 
        \mathcal{C}^{\rm{v2l}}*g_2, \mathcal{R}^{\rm{v2l}}*g_3]),
	\end{aligned} 
\end{equation}
in which $\mathcal{B}$ is of size $\rm{H/32\times W/32}\times 4\rm{(H/32)(W/32)}$. Then it is compressed to $\rm{C}_4^{vis}$ via $Conv\_BN\_ReLU$. The compressed product is finally fused to $\mathcal{V}_4$, denoted as $\mathcal{F}^{\rm{cross}}_4$. 

\subsection{Cross-scale Abstract Semantic Guided Decoder}
\begin{figure}[tbp]
	\centering
	\includegraphics[scale=0.49]{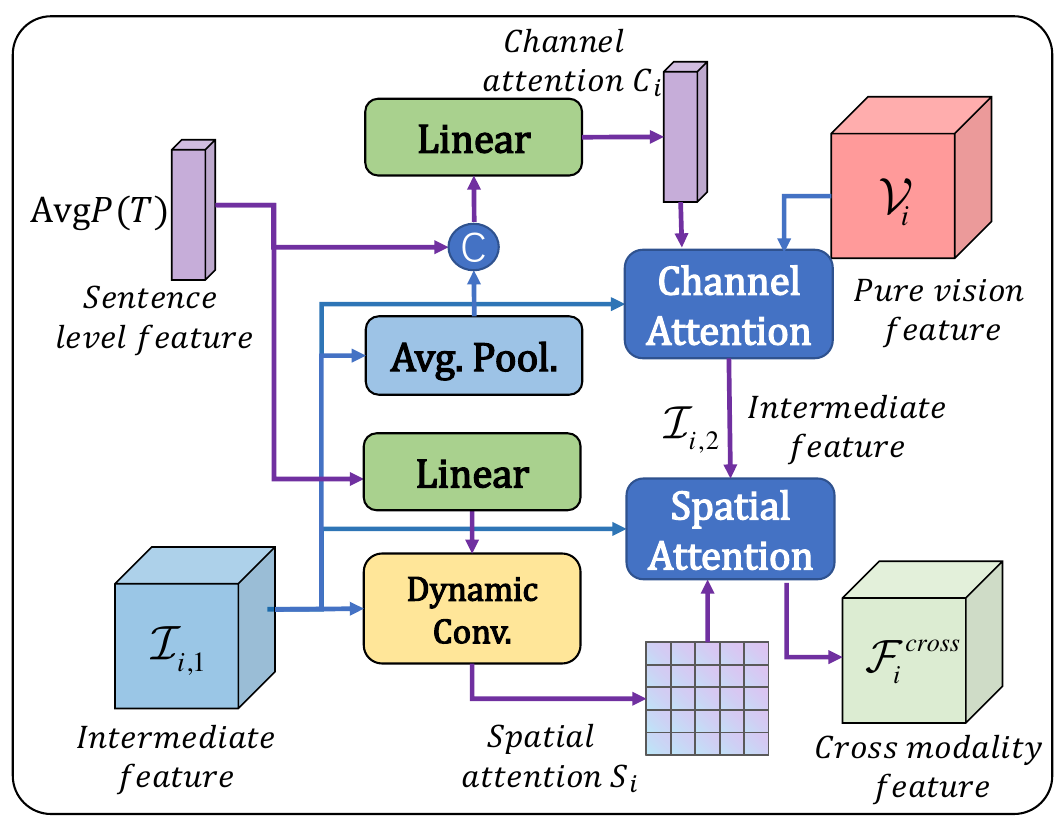}
	\caption{The detailed $i$th stage of CASG. In the process, sentence-level language feature and the semantic-rich feature from previous decoding stages are used to help the pure vision feature supplement effective details of referent.}
	\label{Decoder}
\end{figure}

Integrating multiscale features is beneficial to the prediction of referrent mask with various distributions. On top of the vision and language extractors, S\textsuperscript{2}RM adaptively generates effective deep vision-language features. Extending S\textsuperscript{2}RM to the encoder stages may be a seemingly natural solution to achieving multilevel feature fusion. However, applying nonlocal and correlation based fusion in the early extracting process will invite surging of computation overhead. Instead we propose cross-scale abstract semantic guided decoder (CASG), which allows the fusion to happen in a low-cost way and meantime achieves satisfactory performance. The CASG will use abstract semantic from the upstreaming decoding stages and sentence-level language semantic to generate channel and spatial attention matrices to supplement effective details of the language-relevant area and to weaken the activation of the non-referent region (details see Fig. \ref{Decoder}). 

To be specific, the feature $\mathcal{V}_1$, $\mathcal{V}_2$, $\mathcal{V}_3$, $\mathcal{F}_4^{\rm{cross}}$ and $AvgP(T)$ are together sent to CASG. The process of calculating the fused feature $\mathcal{F}^{\rm{cross}}_i$ of size $\rm{H/2^{\it{i}+\rm{1}}}\times \rm{W/2^{\it{i}+\rm{1}}}\times \rm{C_{\it{i}}^{vis}}$, located in the $i$th to last decoding stage, first requires to generate the channel and spatial attentions as follows:
\begin{equation} \label{add1}
	\begin{aligned}
        & \mathcal{I}_{i,1} = Coord(\mathcal{F}_4^{\rm{cross}},..,\mathcal{F}_{i+1}^{\rm{cross}}) \\
        & C_i = Tanh(MLP(Cat(AvgP(T), AvgP(\mathcal{I}_{i,1})))) \\
        & S_i  = Tanh(\mathcal{I}_{i,1}\otimes Insert(AvgP(T))),i=3,2,1
	\end{aligned} 
\end{equation}
in which $Coord$ represents to re-combine the feature maps from the previous decoding stages, detailed in Eq. \eqref{3th}-Eq. \eqref{1th} for the 3$th$-1$th$ stage; $MLP$ represents a two-layer perceptron; $Tanh$ represents a hyperbolic tangent function; $Insert$ represents to insert a new axis into the first and last dimensions. The temporary tensor $\mathcal{I}_{i,1}$ of size $\rm{H/2^{\it{i}+\rm{1}}\times W/2^{\it{i}+\rm{1}}\times C_{\it{i}}^{vis}}$ is used to assist the generation of channel and spatial attention weights $C_i$ and $S_i$, which are of size $\rm{C_{\it{i}}^{vis}}$ and $\rm{H/2^{\it{i}+\rm{1}}\times W/2^{\it{i}+\rm{1}}\times 1}$. Note that the generation of $S_i$ corresponds to dynamic convolution with activation function $Tanh$. The $\mathcal{F}_i^{\rm{cross}}$ finally is obtained via:
\begin{equation} \label{add2}
	\begin{aligned}
        &\mathcal{I}_{i,2} = Conv(Cat(\mathcal{V}_i*Insert(C_i),\mathcal{I}_{i,1})), \\
        &\mathcal{F}^{\rm{cross}}_i  = Conv(Cat(\mathcal{I}_{i,2}*S_i,\mathcal{I}_{i,1})),
	\end{aligned} 
\end{equation}
in which two skip connections are applied and the feature $\mathcal{V}_i$ is fused into the main branch. The $Insert$ represents to create two new axis in the first two dimensions. The feature $\mathcal{F}^{\rm{cross}}_i$ in this stage is of size $\rm{H/2^{\it{i}+\rm{1}}\times W/2^{\it{i}+\rm{1}}\times C_{\it{i}}^{vis}}$.

In \textbf{the third to last} decoding stage, the intermediate feature $\mathcal{I}_{i,2}$ can be generated in the following manner:
\begin{equation} \label{3th}
	\begin{aligned}
        \mathcal{I}_{3,1} &= Conv(Up2(\mathcal{F}^{\rm{cross}}_4)),
	\end{aligned} 
\end{equation}
in which $Up2$ represents $2\times$ bilinear upsampling. The $\mathcal{F}^{\rm{cross}}_4$ and $\mathcal{F}^{\rm{cross}}_3$ will be used in \textbf{the second to last} decoding stage to generate $\mathcal{I}_{2,1}$, as follows:
\begin{equation} \label{2th}
	\begin{aligned}
        \mathcal{I}_{2,1} &= Conv(Cat(Up4(\mathcal{F}^{\rm{cross}}_4),Up2(\mathcal{F}^{\rm{cross}}_3))),
	\end{aligned} 
\end{equation}
in which the manipulation $Up4$ represents $4\times$ bilinear upsampling manipulation. Similarly, the intermediate feature $\mathcal{I}_{1,1}$ in \textbf{the last stage} is generated in the following manner:
\begin{equation} \label{1th}
	\begin{aligned}
        \mathcal{I}_{1,1} = Conv(Cat(Up8&(\mathcal{F}^{\rm{cross}}_4),\\ &Up4(\mathcal{F}^{\rm{cross}}_3), 
        Up2(\mathcal{F}^{\rm{cross}}_2))),
	\end{aligned} 
\end{equation}
in which the manipulation $Up8$ represents $8\times$ bilinear upsampling manipulation. The final feature $\mathcal{F}^{\rm{cross}}_1$ is fed to another 2D convolution manipulation followed by $4\times$ bilinear upsampling and sigmoid to generate final mask.
%-----------------
\begin{table*}[t]
   \centering
   \resizebox{2.05\columnwidth}{!}{
   \setlength{\tabcolsep}{2mm}{\begin{tabular}{l|l||c|c|c|c|c|c|c|c|c|c}
      \toprule[1pt]
      &\multirow{2}{*}{Method} &
      \multicolumn{3}{c|}{RefCOCO}  & \multicolumn{3}{c|}{RefCOCO+} & \multicolumn{3}{c|}{RefCOCOg}& \multicolumn{1}{c}{ReferIt} \\
      \cline{3-12}
      & & val   & test A & test B & val & test A & test B & val-U & test-U & val-G & test \\
      \hline
      \multirow{10}{*}{\rotatebox[origin=c]{90}{Mean IoU}}
      &DMN$_{18}$~\cite{margffoy2018dynamic}   & 49.78 & 54.83 & 45.13 & 38.88 & 44.22 & 32.29 & -      & -     & 36.76 & 52.81\\
      &MCN$_{20}$~\cite{luo2020multi}          & 62.44 & 64.20 & 59.71 & 50.62 & 54.99 & 44.69 & 49.22 & 49.40 & -  & -    \\
      &CGAN$_{20}$~\cite{luo2020cascade}       & 64.86 & 68.04 & 62.07 & 51.03 & 55.51 & 44.06 & 51.01 & 51.69 & 46.54 & - \\
      &LTS$_{21}$~\cite{lts}        & 65.43 & 67.76 & 63.08 & 54.21 & 58.32 & 48.02 & 54.40 & 54.25 & - & - \\
      &VLT$_{21}$~\cite{vlt}        & 65.65 & 68.29 & 62.73 & 55.50 & 59.20 & 49.36 & 52.99 & 56.65 & 49.76 & - \\
      
      &RefTrans$_{21}$~\cite{li2021referring}         & 74.34 & 76.77 & 70.87 & 66.75 & 70.58 & 59.40 & 66.63 & 67.39 & -  & - \\
      &LAVT$_{22}$~\cite{yang2022lavt}         & 74.46 & 76.89 & 70.94 & 65.81 & 70.97 & 59.23 & 63.34 & 63.62 & 63.66  & - \\
      &Polyfrormer-B$_{23}^{Joint}$~\cite{liu2023polyformer}         & 75.96 & 77.09 & 73.22 & 70.65 & 74.51 & 64.64 & 69.36 & 69.88 & - & - \\
      &M3Att $_{23}$~\cite{liu2023multi}         & 73.60 & 76.23 & 70.36 & 65.34 & 70.50 & 56.98 & 64.92 & 67.37 & 63.90  & - \\
      &SADLR$_{23}$~\cite{yang2023semantics}         & 76.52 & 77.98 & 73.49 & 68.94 & 72.71 & 61.10 & 67.47 & 67.73 & 65.21  & - \\
      \cline{2-12}
      \rule{0pt}{10pt} 
      &\textbf{SwinT-Ours} &74.94 &76.43 &71.22 &66.44 &70.15 &58.89 &65.15 &65.76 &64.21 &66.45 \\
      &\textbf{SwinB-Ours} &76.88 &78.43 &74.01 &70.01 &73.81 &63.21 &69.02 &68.49 &66.63 &68.46\\
      \midrule[1pt]
      \multirow{18}{*}{\rotatebox[origin=c]{90}{Overall IoU}}
      &RMI+DCRF$_{17}$~\cite{liu2017recurrent}      &45.18 &45.69 &45.57 &29.86 &30.48 &29.50  &- &- &34.52 & 58.73 \\
      &KWA$_{18}$~\cite{shi2018key}                 &-     &-     &-     &-     &-     &- &- &- &36.92 &59.19 \\
      &RRN$_{18}$~\cite{li2018referring}       & 55.33 & 57.26 & 53.93 & 39.75 & 42.15 & 36.11 & -     & -     & 36.45 & 63.63 \\
      &MAttNet$_{18}$~\cite{yu2018mattnet}     & 56.51 & 62.37 & 51.70 & 46.67 & 52.39 & 40.08 & 47.64 & 48.61 & - & -     \\
      &STEP$_{19}$~\cite{chen2019see}               &60.04 &63.46 &57.97 &48.19 &52.33 &40.41 &- &- &46.40 &64.13\\
      &CMSA+DCRF$_{19}$~\cite{ye2019cross}          & 58.32 & 60.61 & 55.09 & 43.76 & 47.60 & 37.89 & -     & -     & 39.98 & 63.80 \\
      &CMPC+DCRF$_{20}$~\cite{huang2020referring}   & 61.36 & 64.53 & 59.64 & 49.56 & 53.44 & 43.23 & -     & -     & 49.05 & 65.53 \\
      &LSCM+DCRF$_{20}$~\cite{hui2020linguistic}    & 61.47 & 64.99 & 59.55 & 49.34 & 53.12 & 43.50 & -     & -     & 48.05 & 66.57  \\
      &SANet$_{21}$~\cite{lin2021structured}                             &61.84 &64.95 &57.43 &50.38 &55.36 &42.74  &- &- &44.53 &65.88\\
      &BRINet+DCRF$_{20}$~\cite{hu2020bi123}           &61.35 &63.37 &59.57 &48.57 &52.87 &42.13 &48.04 &- &- & 63.46 \\
      &CEFNet$_{21}$~\cite{feng2021encoder}    & 62.76 & 65.69 & 59.67 & 51.50 & 55.24 & 43.01  & - & -  & 51.93 & 66.70 \\ 
      % &BUSNet     & 63.27 & 66.41 & 61.39 & 51.76 & 56.87 & 44.13 & - & - & 50.56 & -  \\
      &ReSTR$_{22}$~\cite{Kim2022ReSTRCR}        & 67.22 & 69.30 & 64.45 & 55.78 & 60.44 & 48.27  & -     &  -   & 54.48 & -   \\
      &ISPNet$_{22}$~\cite{Liu2022InstanceSpecificFP}      &65.19 &68.45 &62.73 &52.70 &56.77 &46.39 &53.00 &50.08 &52.39 &-\\
      &CRIS$_{22}$~\cite{wang2022cris}         & 70.47 & 73.18 & 66.10 & 62.27 & 68.08 & 53.68 & 59.87 & 60.36 & -  & - \\
      &LAVT$_{22}$~\cite{yang2022lavt}         & 72.73 & 75.82 & 68.79 & 62.14 & 68.38 & 55.10 & 61.24 & 62.09 & 60.50 & - \\
      &Polyfrormer-B$_{23}^{Joint}$~\cite{liu2023polyformer}         & 74.82 & 76.64 & 71.06 & 67.64 & 72.89 & 59.33 & 67.76 & 69.05 & - & - \\
      &FSFINet$_{23}$~\cite{yang2023referring}                             &71.23 &74.34 &68.31 &60.84 &66.49 &53.24 &61.51 &61.78 &60.25 &73.36\\
      &SADLR$_{23}$ ~\cite{yang2023semantics}         & 74.24 & 76.25 & 70.06 & 64.28 & 69.09 & 55.19 & 63.60 & 63.56 & 61.16 & - \\
      \cline{2-12}
      \rule{0pt}{10pt}
      &\textbf{SwinT-Ours} &72.08 &74.15 &67.45 &61.73 &65.90 &52.92  &61.77 &61.76 &60.44 &72.31\\
      &\textbf{SwinB-Ours} &74.35 &76.57 &70.44 &65.39 &70.63 &57.33 &65.37 &65.30  &62.64 &73.88\\
      \bottomrule[1pt]
   \end{tabular}}}
   \caption{Quantitative comparison of our method and other state-of-the-art algorithms. Mean IoU and Overall IoU are adopted to measure the performance. The 'DCRF' represents the post-processing step using Dense CRF.  The ${}^{joint}$ denotes the only results trained on \textbf{joint} of RefCOCO/+/g and Flicker30K~\cite{plummer2015flickr30k}. The subscript represents their publication years.}
   \label{tab:1}
\end{table*}

\begin{figure*}[htbp]
	\centering
	\includegraphics[scale=0.56]{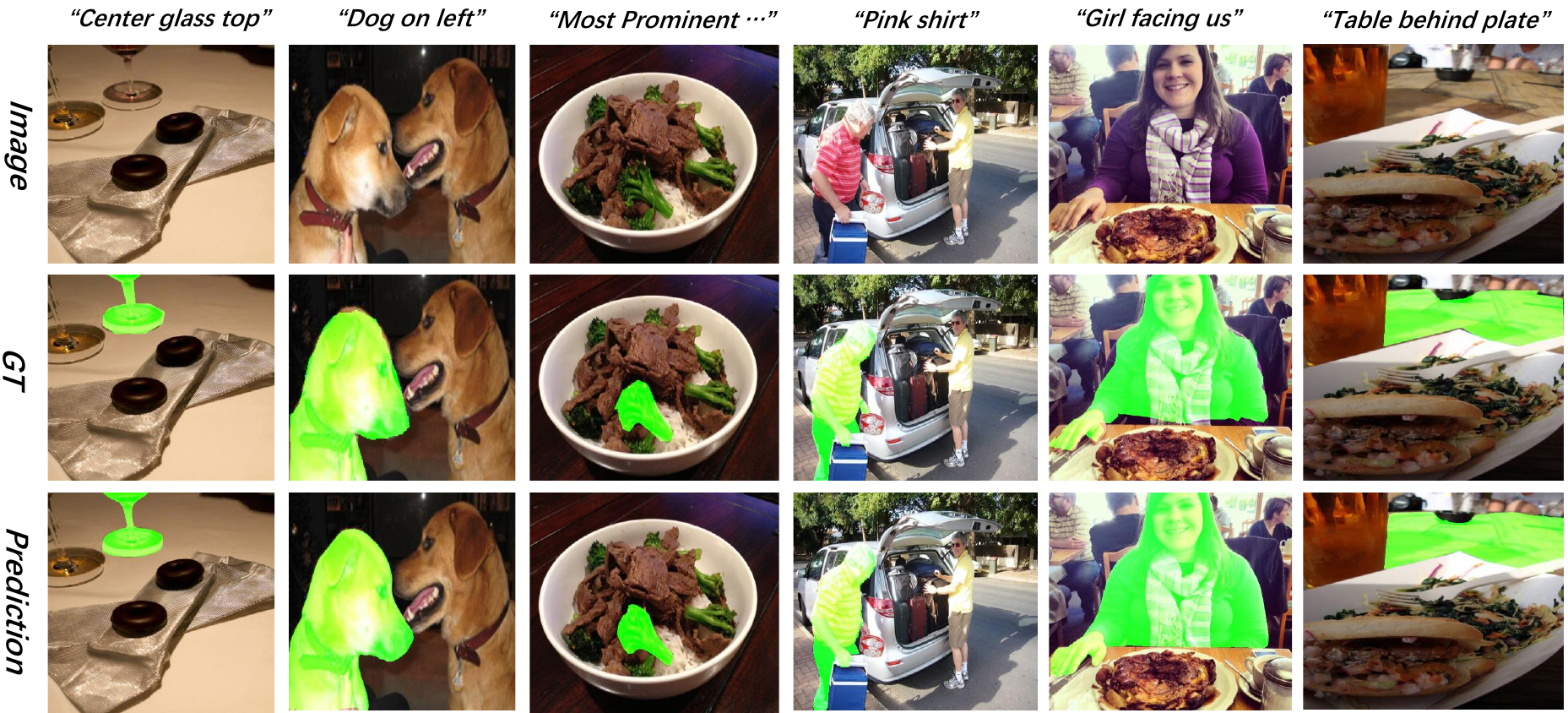}
	\caption{Visualization results of proposed techniques on some samples selected from the validation set of RefCOCO.}
	\label{Visualization}
\end{figure*}

\section{Experiments}
\label{studies}
\subsection{Datasets and Evaluation Metrics}
\textbf{Datasets:} In this work, we benchmark on four current challenging datasets, namely RefCOCO (UNC)~\cite{yu2016modeling}, RefCOCO+(UNC+)~\cite{yu2016modeling}, RefCOCOg ~\cite{mao2016generation}, ReferIt~\cite{kazemzadeh2014referitgame}.

1) The RefCOCO dataset is formed by picking up the samples from MSCOCO dataset resorting to a two-player games. It totally contains 19994 images and 142209 language expressions, which, to sum up, have average length of 3.5 words and describe 50000 regions. In general, the dataset uses 120624, 10834, 5657, and 5095 image-language pairs in training, validating, testing in set A, and testing in set B.

2) Similar to RefCOCO, RefCOCO+ is also selected from MSCOCO with 141564 expressions describing 49856 regions in 19992 images. Different from RefCOCO, the expression within tends to focus on the appearance of the referred regions, less including their location information. The dataset divides 120191, 10758, 5726, and 4889 image-expression pairs into training part, validation part, A part, and B part.

3) The RefCOCOg is still selected from MSCOCO dataset but via Amazon Mechanical Turk. The dataset has 104560 language expressions to refer to 54822 regions in 26711 images. Compared to the expression in RefCOCO, its language is longer and more complex (describe both appearance and distribution of the referent), with average length of 8.4 words. The dataset has UMD and Google patterns. 

4) The ReferIt is collected from SALAPR-12 dataset, containing 130525  expressions referring to 96654 regions across 19894 images. We obey the berkeley partition pattern dividing the cleaned samples into training set of 54127 pairs, testing set of 5842 pairs, and validation set of 60103 pairs, respectively. Usually the expression is more succint.

\textbf{Metrics:} In the following experiments, three widely-accepted metrics are used to benchmark our proposed techniques, namely overall Intersection-Over-Union (OIoU), mean Intersection-Over-Union (MIoU), and Precision@$\rm X$ ($\rm P$@$\rm X$). The OIoU will calculate ratio of the total intersection region over total union region for the ground truth and their prediction across all test samples, of which each is composed of a image and a related expression. The MIoU calculates the average value of IoU for all test samples. The $\rm Prec$@$\rm X$ reports the percentage of IoU score exceeding the specified threshold value $\rm X$ in the test set, which focuses on estimating the targeting ability of the method. Following previous works, the thresholding value $\rm X$ is selected from the set of $\{0.5, 0.6, 0.7, 0.8, 0.9\}$.
\begin{table}[tbp]
	\centering
	\caption{The scores of P@0.5, P@0.6, P@0.7, P@0.8, P@0.9 over the RefCOCO validation set and Flops.}
	\resizebox{0.5\textwidth}{!}{
		\setlength\tabcolsep{2.5pt}
		\renewcommand\arraystretch{1}
		\begin{tabular}{l||c|c|c|c|c|c}
			\toprule[1pt]
			Method  &P@0.5$\uparrow$ &P@0.6$\uparrow$ &P@0.7$\uparrow$ &P@0.8$\uparrow$  &P@0.9$\uparrow$ & Flops$\downarrow$\\
			% 			\hline
			% 			\hline
			\hline
			% LSCM~\cite{hui2020linguistic}  &70.84 &63.82 &53.67 &38.69 &12.06 \\
			% CMPC~\cite{huang2020referring} &71.27 &64.44 &55.03 //&39.28 &12.89 \\
			% MCN~\cite{luo2020multi}  &76.60 &70.33 &58.39 &33.68 &5.26 \\
                BRINet~\cite{hu2020bi123} &71.83 &65.05 &55.64 &39.36 &11.21, &367.63 \\
                LTS~\cite{lts}  &75.16 &69.51 &60.74 &45.17 &14.41 &133.3G\\
                CEFNet~\cite{feng2021encoder} &73.95 &69.58 &62.59 &49.61 &20.63&112.92 \\
			VLT~\cite{vlt} &76.20 &- &- &- & -& 142.6G\\
                LAVT~\cite{yang2022lavt} &84.46	&80.90	&75.28 &64.70 &34.30 &197.4G \\
                M3Att~\cite{liu2023multi} &79.01	&74.94	&68.16 &51.21 &17.70 & - \\
                SADLR~\cite{yang2023semantics} &86.90 &83.68 &78.76 &67.93 &37.36 & 203.5G \\
                \hline
                SwinT-Ours &85.39 &81.58 &75.52 &63.87 &33.16 &92.9G\\
                SwinB-Ours &87.21 &84.11 &79.20 &68.64 &37.18 &199.0G\\
			\bottomrule[1pt]
	\end{tabular}}	
	\label{tab:PX}
\end{table} 

\begin{table}[htbp]
	\vspace{0mm}
	\setlength{\tabcolsep}{6pt}
	\centering
	\caption{OIoU comparison on the validation partition of RefCOCO, RefCOCO+, RefCOCOg, and testing part of ReferiIt against expressions of different length.} 
	%\begin{center}
	\renewcommand{\arraystretch}{1.5}
	\resizebox{0.43\textwidth}{!}{
		\begin{tabular}{p{0.5cm}<{\centering}|p{2.0cm}<{\centering}||p{0.6cm}<{\centering}|p{0.6cm}<{\centering}|p{0.6cm}<{\centering}|p{0.6cm}<{\centering}}
			\toprule[1pt]
			\multirow{1}{*}{} &Length &1-2 &3 &4-5 &6-20 \\
			\hline
			\multirow{5}{*}{\rotatebox[origin=c]{90}{RefCOCO}}
			% &R+LSTM~\cite{liu2017recurrent}    &43.66 &40.60 &33.98 &24.91 \\
			&R+RMI~\cite{liu2017recurrent}     &44.51 &41.86 &35.05 &25.95 \\
			&BRINet~\cite{hu2020bi123}            &65.99 &64.83 &56.97 &45.65 \\
			% &VCM~\cite{feng2021encoder}    &68.18 &66.14 &56.82 &46.01 \\
			&ACM~\cite{feng2021encoder}    &68.73 &65.58 &57.32 &45.90 \\
			&ReSTR~\cite{Kim2022ReSTRCR}  &72.38 &69.46 &61.19 &50.21 \\
                &FSFI-R101~\cite{yang2023referring} &76.10 &72.51 &66.51 &58.39 \\
                \hline
			&SwinT-Ours                    &77.57 &75.15 &68.50 &60.20 \\
                &SwinB-Ours                    &79.67 &76.50 &71.15 &63.57 \\
			\bottomrule[1pt]
	\end{tabular}}
	
	\vspace{3mm}
	\resizebox{0.43\textwidth}{!}{
		\begin{tabular}{p{0.5cm}<{\centering}|p{2.0cm}<{\centering}||p{0.6cm}<{\centering}|p{0.6cm}<{\centering}|p{0.6cm}<{\centering}|p{0.7cm}<{\centering}}
			\toprule[1pt]
			\multirow{1}{*}{} &Length &1-2 &3 &4-5 &6-20 \\
			\hline
			\multirow{5}{*}{\rotatebox[origin=c]{90}{RefCOCO+}}
			% &R+LSTM~\cite{liu2017recurrent}    &34.40 &24.04 &19.31 &12.30 \\
			&R+RMI~\cite{liu2017recurrent}     &35.72 &25.41 &21.73 &14.37 \\
			&BRINet~\cite{hu2020bi123}            &59.12 &46.89 &40.57 &31.32 \\
			% &VCM~\cite{feng2021encoder}     &60.87 &48.88 &43.79 &29.45 \\
			&ACM~\cite{feng2021encoder}     &61.62 &52.18 &43.46 &31.52 \\
			&ReSTR~\cite{Kim2022ReSTRCR}                         &65.72 &54.81 &47.65 &37.02\\
                &FSFI-R101~\cite{yang2023referring}  &69.85 &59.34 &52.76 &40.72 \\
                \hline
			&SwinT-Ours                         &71.79 &64.15 &56.29 &46.01\\
                &SwinB-Ours                         &75.09 &66.43 &60.82 &50.87\\
			\bottomrule[1pt]
	\end{tabular}}
	
	\vspace{3mm}
	\resizebox{0.43\textwidth}{!}{
		\begin{tabular}{p{0.5cm}<{\centering}|p{2.0cm}<{\centering}||p{0.6cm}<{\centering}|p{0.6cm}<{\centering}|p{0.6cm}<{\centering}|p{0.7cm}<{\centering}}
			\toprule[1pt]
			\multirow{1}{*}{} &Length &1-5 &6-7 &8-10 &11-20 \\
			\hline
			\multirow{5}{*}{\rotatebox[origin=c]{90}{RefCOCOg (Google)}}
			% &R+LSTM~\cite{liu2017recurrent}    &32.29 &28.27 &27.33 &26.61 \\
			&R+RMI~\cite{liu2017recurrent}     &35.34 &31.76 &30.66 &30.56 \\
			&BRINet~\cite{hu2020bi123}            &51.93 &47.55 &46.33 &46.49 \\
			% &VCM~\cite{feng2021encoder}       &57.96 &52.19 &48.78 &46.67\\
			&ACM~\cite{feng2021encoder}       &59.92 &52.94 &49.56 &46.21\\
			&ReSTR~\cite{Kim2022ReSTRCR}    
			&58.72 &53.43 &53.96 &51.91\\
                &FSFI-R101~\cite{yang2023referring} &64.34 &59.82 &55.25 &52.52 \\
                \hline
                &SwinT-Ours                       &64.15 &62.46 &57.86 &59.44\\
			&SwinB-Ours                       &66.51 &65.02 &59.03 &62.16\\
			\bottomrule[1pt]
	\end{tabular}}

        \vspace{3mm}
	\resizebox{0.43\textwidth}{!}{
		\begin{tabular}{p{0.5cm}<{\centering}|p{2.0cm}<{\centering}||p{0.6cm}<{\centering}|p{0.6cm}<{\centering}|p{0.6cm}<{\centering}|p{0.7cm}<{\centering}}
			\toprule[1pt]
			\multirow{1}{*}{} &Length &1 &2 &3-4 &5-20 \\
			\hline 
			\multirow{5}{*}{\rotatebox[origin=c]{90}{ReferIt}}
			&R+RMI~\cite{liu2017recurrent}     &68.11 &52.73 &45.69 &34.53 \\
			&BRINet~\cite{hu2020bi123}            &75.28 &62.62 &56.14 &44.40 \\
			% &VCM~\cite{feng2021encoder}       &77.73 &66.02 &59.74 &45.75\\
			&ACM~\cite{feng2021encoder}       &78.19 &66.63 &60.30 &46.18\\
			&ReSTR~\cite{Kim2022ReSTRCR}    
			&80.82 &69.78 &63.66 &50.73\\
                &FSFI-R101~\cite{yang2023referring} &83.10 &71.42 &64.86 &54.93 \\
                \hline
                &SwinT-Ours                       &83.42 &71.66 &66.01 &57.08\\
			&SwinB-Ours                       &84.36 &73.70 &67.82 &59.34\\
			\bottomrule[1pt]
	\end{tabular}}
	\label{tab:3}
\end{table}
%%%%%%%%%%%%%%%%%%%%%%%%%%%%%%%%%%%%%%%%%%%%%%%%%%%%%%%%%%%%%%%%%%%%%%%%%%%%%%%%%%%%%%%%%%%%%%%%%%%%%%%%%%%%%%%
\subsection{Implementation Details}
\textbf{Experimental settings.} Following previous SOTA method Polyformer~\cite{liu2023polyformer}, the original image is resized to $512\times 512$ before fed into the network for all of the datasets. All of the involved experiments, based on Pytorch and HuggingFace framewok, are running on a workstation with i7-10700 core, 32G DRAM, and an RTX 3090 GPU card. The pretrained weights of Swin transformer are generated in the classification task of ImageNet22K. Other parameters are initialized randomly. The 12-layer Bert with hidden dimension of $768$ is initialized with the HuggingFace provided weights. The clipping length of the language is set to 20. We use AdamW optimizer with weight decay 0.01, an initial learning rate 0.00005, and polynomial learning rate decay strategy to drive the parameter updating process. We use dice loss to supervise the output of the model, and the batch size is set to 32. After 40 epochs, the training process will be terminated. To speed up the whole training process, we accept the same sampling strategy in LAVT and SADLR, in which image and one random sampling from its bundled expressions will be used only once in each epoch. Besides, to further conserve resources, the ablation studies are run on Swin tiny.

%%%%%%%%%%%%%%%%%%%%%%%%%%%%%%%%%%%%%%%%%%%%%%%%%%%%%%%%%%%%%%%%%%%%%%%%%%%%%%%%%%%%%%%%%%%%%%%%%%%%%%%%%%%%%%%
%%%%%%%%%%%%%%%%%%%%%%%%%%%%%%%%%%%%%%%%%%%%%%%%%%%%%%%%%%%%%%%%%%%%%%%%%%%%%%%%%%%%%%%%%%%%%%%%%%%%%%%%%%%%%%%

% \begin{figure*}[htbp]
% 	\centering
% 	\includegraphics[scale=0.58]{figs/Visualization.pdf}
% 	\caption{Visualization results of proposed techniques. The samples are also selected from the validation set of UNC.}
% 	\label{Visualization}
% \end{figure*}

\begin{figure*}[htbp]
	\centering
	\includegraphics[scale=0.41]{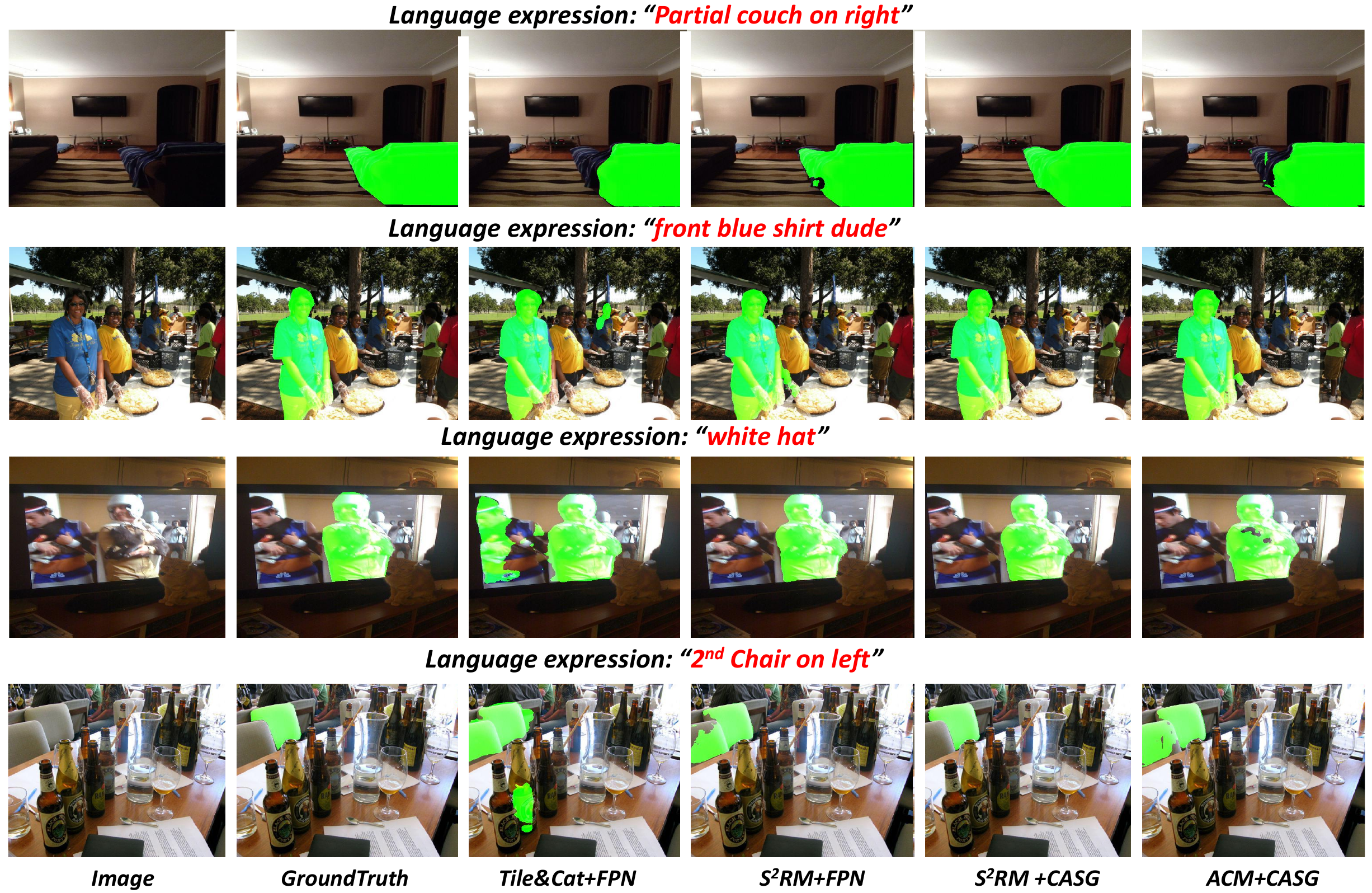}
	\caption{Visualization of the proposed method. The samples are selected from the validation set of RefCOCO.} 
	\label{fig:add}
\end{figure*}

\subsection{Comparison with the SOTAs}
To demonstrate overall performance of our method, we document its OIoU and MIoU scores in Tab. \ref{tab:1}, as well as the results from other state-of-the-art methods. The involved algorithms include RMI~\cite{liu2017recurrent},  DMN~\cite{margffoy2018dynamic}, KWA~\cite{shi2018key},  RRN\cite{li2018referring}, MATTNet~\cite{yu2018mattnet},  CMSA~\cite{huang2020referring}, STEP~\cite{chen2019see}, CGAN~\cite{luo2020cascade}, BRINet~\cite{hu2020bi123}, LSCM~\cite{hui2020linguistic}, CMPC~\cite{huang2020referring}, CEFNet~\cite{feng2021encoder}, LTS~\cite{lts}, VLT~\cite{vlt}, RefTrans~\cite{li2021referring}, ISPNet~\cite{Liu2022InstanceSpecificFP}, and ReSTR~\cite{Kim2022ReSTRCR}, LAVT~\cite{yang2022lavt}, Polyformer~\cite{liu2023polyformer}, M3Att~\cite{liu2023multi}, SADLR~\cite{yang2023semantics}. The results other than the last two rows in each sub-table are borrowed from their respective official publications. The SwinT-Ours represents to use Swin Tiny to extract vision features, to apply 12-Layer Bert to extract the language feature, to accept S\textsuperscript{2}RM to interact the features from the two modalities, and to decode mask of referent using CASG decoder. Different from SwinT-Ours, SwinB-Ours resorts to Swin Base to extract vision feature. Note that in the G-Ref entry, the U-val, U-test, and G-val represent validation part of UMD partition, test part of UMD partition, and validation part of goolge partition, respectively. Following M3Att, the P@X score and Flops of our method and other state of the art algorithms are shown in Tab. \ref{tab:PX}.

From Tab. \ref{tab:1}, it can be seen that our network can generate good performance on the four datasets. For SwinT-Ours, it surpasses the previous method VLT by 8.64, 10.47, and 11.90 points on average against mIoU on RefCOCO, RefCOCO+, and RefCOCOg, respectively. The performance of SwinT-Ours is on par with RefTrans, which is in addition pretrained on large-scale dataset Visual Genome~\cite{krishna2017visual}. In comparison with pure-transformer based ReSTR, SwinT-Ours on oIoU achieves average improvements of 4.24, 5.35, and 5.96 points on the three datasets. 
For SwinB-Ours, it can consistently outperform other state of the art methods. Compared to the early fusion method CEFNet and LAVT, SwinB-Ours can generate 11.79 and 2.31 points improvement on averaged oIoU score across all involved datasets, respectively. In comparison with the current front-runner SADLR, SwinB-Ours has average 1.17 and
1.04 improved performance on oIoU and  mIoU across all three datasets, respectively. To our surprise, compared to Polyformer-B$^{Joint}$ pretrained on far more larger joint of datasets (RefCOCO/+/g, Flicker 30K entities \cite{plummer2015flickr30k}), SwinB-ours approaches its scores, especially on mIoU.
Note that LAVT, SADLR, and Polyformer-B all use Swin base and Bert Base to extract vision and language features. 

From Tab. \ref{tab:PX}, it can be seen that SwinT-Ours (Swin tiny) surpasses LAVT (Swin Base) on P@0.5, P@0.6, P@0.7, and GFlops by impressive margin of 0.93, 0.68, and 0.24 point, and 104.5G, respectively. Across different P@X, SwinB-Ours achieves on average 3.34 superior performance over LAVT. Compared to the first-localization-then-segment method LTS, SwinT-Ours and SwinB-Ours have 14.91 and 18.27 average gains, respectively. These facts show that our network owns a strong ability in capturing the relationship of image content and language. The statistical results of our method on the four datasets against different language lengths, as well as those from some other typical advanced works, are shown in Tab. \ref{tab:3}. The SwinT-Ours can surpass FSFI-R101 across datasets given expressions with varying length. The consistent improvements of our method over other SOTA methods validate the robustness of our proposed techniques. Note that different from LAVT, FSFINet, and SADLR, \textbf{our network does not need to fuse language information into the early extracting stages. Therefore, for image with multiple expressions that is usual in the involved datasets, our network free from extra feature extraction should have more less computation during inference.}

In Fig. \ref{Visualization}, we show some samples of SwinT-Ours on the validation partition part of RefCOCO. Across all samples, it can be seen that our method can generate more fine boundary than the annotation, for example, the missing patch of left dog head is retrieved in column of "dog on left". In "center glass top" and "girl facing us" samples, the boundary of predicted referents are more delicate.

\begin{table}[tbp]
	\centering
	\caption{Ablation studies on the validation set of RefCOCO. The details of S\textsuperscript{2}RM are investigated.}
	\resizebox{0.5\textwidth}{!}{
		\setlength\tabcolsep{3pt}
		\renewcommand\arraystretch{1.45}
		\begin{tabular}{l||c|c|c|c|c}
			\toprule
			Method &oIoU$\uparrow$  &mIoU$\uparrow$ &P@0.5$\uparrow$  &P@0.7$\uparrow$ &P@0.9$\uparrow$\\
			\hline
            \textcolor{red}{W.} Tile \& Concat. &60.55 &61.45 &70.21 &52.93 &12.34\\
            % \textcolor{red}{W.} Tile \& Nonlocal &65.79 &67.39 &77.04 &62.47 &19.51\\
            \textcolor{red}{W.} S\textsuperscript{2}RM (-Shift-l2v) &65.67 &66.67 &76.97 &61.26 &17.05\\
            \textcolor{red}{W.} S\textsuperscript{2}RM (-Shift-v2l) &67.18 &67.63 &80.51 &62.96 &11.85\\
            \textcolor{red}{W.} S\textsuperscript{2}RM (-Shift) &68.16 &69.27 &80.41 &65.35 &19.53\\
            \textcolor{red}{W.} S\textsuperscript{2}RM (-l2v) &66.19 &67.81 &78.03 &62.74 &18.15\\
            \textcolor{red}{W.} S\textsuperscript{2}RM (-v2l) &68.96 &70.36 &81.35 4 &67.69  &21.02\\
            \textcolor{red}{W.} S\textsuperscript{2}RM (-Balance) &  69.33 &70.51 &76.14 &67.90 &21.40\\
            \textcolor{red}{W.} S\textsuperscript{2}RM (Full)  & 69.57 &71.86 &83.37 &71.49 &23.07\\
            \bottomrule
            \end{tabular}}	
	\label{tab:s2RM}
\end{table}

\subsection{Ablation Studies}
We validate effectiveness of the proposed S\textsuperscript{2}RM and CASG on the validation part of RefCOCO, following the LAVT, SADLR, and \emph{etc}. The results are shown in Tab. \ref{tab:s2RM} and \ref{tab:casg}. Symbol \textcolor{red}{W.} is used to represent the abbreviation of word \emph{with}. To ablate the design of S\textsuperscript{2}RM, we resorts to Swin tiny to extract the semantic of image content, Bert base to extract the semantic of expression, and FPN structure to decode the referent appearance. Note that the decoder is borrowed from the literature \cite{lin2017feature}. To validate the design of decoder CASG, we use Swin tiny to extract the semantic of image content, Bert base to extract the semantic of expression, and full-packaged S\textsuperscript{2}RM to fuse the vision and language features.

% \begin{table}[htbp]  
% 	\centering
% 	\caption{The impact of the fusion stages on \textcolor{red}{W.} FSFI (16)+MAED on the UNC validation set.}
% 	\resizebox{0.49\textwidth}{!}{
% 		\setlength\tabcolsep{4pt}
% 		\renewcommand\arraystretch{1.3}
% 		\begin{tabular}{l|c|c|c|c|c}
% 		\toprule
% 			Method &oIoU$\uparrow$ &mIoU$\uparrow$ &P@0.5$\uparrow$  &P@0.7$\uparrow$  &P@0.9$\uparrow$\\
% 			\hline
%             $512\times 512$ &72.08 &74.94 &85.39 &75.52 &33.16\\
%             $480\times 480$ &71.43 &74.40 &84.97 &75.44 &32.15\\
%             $448\times 448$ & 71.45 &74.29 &84.86 &75.38 &31.81\\
%             $384\times 384$ & 70.80 &73.52 &84.31 &74.22 &30.41\\
%             $320\times 320$ & 68.14 &71.00 &82.12 &70.29 &24.31\\
% 		\bottomrule
% 	\end{tabular}}	
% 	\label{tab:fusion}
% \end{table} 

\begin{table}[tbp]
	\centering
	\caption{Ablation studies on the validation set of RefCOCO. The details of CASG are investigated.}
	\resizebox{0.5\textwidth}{!}{
		\setlength\tabcolsep{3pt}
		\renewcommand\arraystretch{1.45}
		\begin{tabular}{l||c|c|c|c|c}
			\toprule
			Method &oIoU$\uparrow$  &mIoU$\uparrow$ &P@0.5$\uparrow$  &P@0.7$\uparrow$ &P@0.9$\uparrow$\\
			\hline
            \textcolor{red}{W.} CASG (-Lang.)& 71.65 &74.59 &84.88 &75.86 &32.79\\
            \textcolor{red}{W.} CASG (-Cha.)& 71.07 &73.53 &83.85  &74.19 &31.90\\
            \textcolor{red}{W.} CASG (-Spa.)& 70.99 &74.14 &84.36 &74.73 &32.00\\
            \textcolor{red}{W.} CASG (Full)& 72.08 &74.94 &85.39 &75.52  &33.16\\
            \textcolor{red}{W.} CASG ($Sig.$)& 71.02 &74.34 &80.82 &75.26 &33.09\\
			\bottomrule
	\end{tabular}}	
	\label{tab:casg}
\end{table}

1) $\emph{Validation of Main Contributions.}$ From Tab. \ref{tab:s2RM}, it can be seen that \textcolor{red}{W.} S\textsuperscript{2}RM+FPN surpasses the baseline method \textcolor{red}{W.} Tile+Concat by 9.02, 10.41, and 15.49 points against oIoU, mIoU, and average arithmatic P@X, respectively. The facts demonstrate that the proposed S\textsuperscript{2}RM is very effective in terms of fusing vision and language features. Different from past works CEFNet~\cite{feng2021encoder} and LAVT~\cite{yang2022lavt} instilling the language semantic into the early stages of vision encoder, CASG resorts to cross-modality features from the previous decoding stages and sentence-level language feature to generate attention feature maps, finally purifying the extracted multi-scale vision features to generate more delicate referent mask. Overall on the RefCOCO validation partition, S\textsuperscript{2}RM coupled with CASG (\textcolor{red}{W.} S\textsuperscript{2}RM (Full) in Tab. \ref{tab:casg}) surpasses S\textsuperscript{2}RM with FPN (\textcolor{red}{W.} S\textsuperscript{2}RM (Full) in Tab. \ref{tab:s2RM}) by 2.51, 3.08, and 5.26 points against oIoU, mIoU, mean arithmatic P@X, respectively.

2) $\emph{Validation of S\textsuperscript{2}RM Details}$. To validate the design of co-parsing pattern, we remove $l2v$ and $v2l$ from shift-uninstalled S\textsuperscript{2}RM and full-packaged S\textsuperscript{2}RM, respectively. Their results are documented in the entries of \textcolor{red}{W.} S\textsuperscript{2}RM (-Shift-l2v), \textcolor{red}{W.} S\textsuperscript{2}RM (-Shift-v2l), \textcolor{red}{W.} S\textsuperscript{2}RM (-l2v), and \textcolor{red}{W.} S\textsuperscript{2}RM (-v2l) in Tab. \ref{tab:s2RM}. On the three metrics, \textcolor{red}{W.} S\textsuperscript{2}RM (-Shift) outperforms \textcolor{red}{W.} S\textsuperscript{2}RM (-Shift-l2v) by 2.49, 2.60, 3.46 points and \textcolor{red}{W.} S\textsuperscript{2}RM (-Shift-v2l) by 0.98, 1.64, 3.59 points, respectively. For \textcolor{red}{W.} S\textsuperscript{2}RM, it surpasses \textcolor{red}{W.} S\textsuperscript{2}RM (-l2v) by 3.38, 4.05, 6.95 points and \textcolor{red}{W.} S\textsuperscript{2}RM (-v2l) by 0.61, 1.50, 2.80 points, respectively. These improvements show the effectiveness of our bi-parsing strategy. To validate the design of balance strategy in S\textsuperscript{2}RM, we remove the third step from S\textsuperscript{2}RM, and the results are shown in item of \textcolor{red}{W.} S\textsuperscript{2}RM (-Balance) in Tab. \ref{tab:s2RM}. It can be seen that on the three metrics, \textcolor{red}{W.} S\textsuperscript{2}RM (Full) has 0.24, 1.35, and 2.51 better performance than \textcolor{red}{W.} S\textsuperscript{2}RM (-Balance). The gains of \textcolor{red}{W.} S\textsuperscript{2}RM (Full) over \textcolor{red}{W.} S\textsuperscript{2}RM (-Shift) demonstrate the effectiveness of shifting mechanism.

3) $\emph{Validation of CASG Details}$.  To validate the design of multiscale feature integration process, we remove the spatial and channel guidance and language integration from the decoder, respectively. Their results are shown in entries of  \textcolor{red}{W.} CASG (-Spa.),  \textcolor{red}{W.} CASG (-Cha.), and \textcolor{red}{W.} CASG (-Lang.) in Tab. \ref{tab:stg}, respectively.
It can be seen that our design is very effective in decoding process of referent mask. To be specific, full-packaged \textcolor{red}{W.} CASG (Full) has 1.01, 1.41, 1.03 improvements over \textcolor{red}{W.} CASG (-Spa.) and 1.09, 0.80, 1.54 improvements over \textcolor{red}{W.} CASG (-Cha.). The \textcolor{red}{W.} CASG (Full) surpasses CASG (-Lang.) by 0.43, 0.35 and 0.18 point on the three metrics, respectively. We also use Sigmoid to replace the adopted $Tanh$ manipulation in the decoding process. The results can be seen in the entry of \textcolor{red}{W.} CASG ($Sig.$). To validate the application of the attention mechanism, we gradually remove the attention-guided decoding stage in reverse order. Their results are shown in Tab. \ref{tab:stg}. It can be seen that current employment of four stages achieves best performance.

4) $\emph{Comparison with Other Fusion Method}$. To further benchmark the performance of our proposed fusion method, which usually plays a core role in referring image segmentation task, we replace S\textsuperscript{2}RM in the designed network with ACM in CEFNet~\cite{feng2021encoder} and PWAM in LAVT~\cite{yang2022lavt}, utilizing FPN and the proposed CASG to decode the mask of referent. The quantitative results are shown in Tab. \ref{tab:cmp_other}. All of the method will use Swin tiny and Bert Base to extract vision and language features. From the table, it can be seen that on FPN, our S\textsuperscript{2}RM (S\textsuperscript{2}RM+FPN) tops ACM (ACM+FPN) by 2.98, 3.39, 4.99 points, and surpasses PWAM (PWAM+FPN) by 1.52, 1.16, 1.71 on oIoU, mIoU, and arithmetic mean P@X, respectively. On the proposed decoder CASG, the respective improvements of our fusion method (S\textsuperscript{2}RM+CASG) over ACM (ACM+CASG) and PWAM (PWAM+CASG) are 2.75, 2.55, 3.30 and 1.21, 1.13, 1.68 points, respectively.

5) $\emph{Visualization Results}$. In Fig. \ref{fig:add}, we show some visualization results, which are selected from the prediction of RefCOCO validation part. It can be seen that our proposed S\textsuperscript{2}RM+CASG+SwinT+BertBase generates more satisfactory results over other combinations.

\subsection{Failure Cases and Future Prospects}
\begin{figure}[tbp]
	\centering
	\includegraphics[scale=0.52]{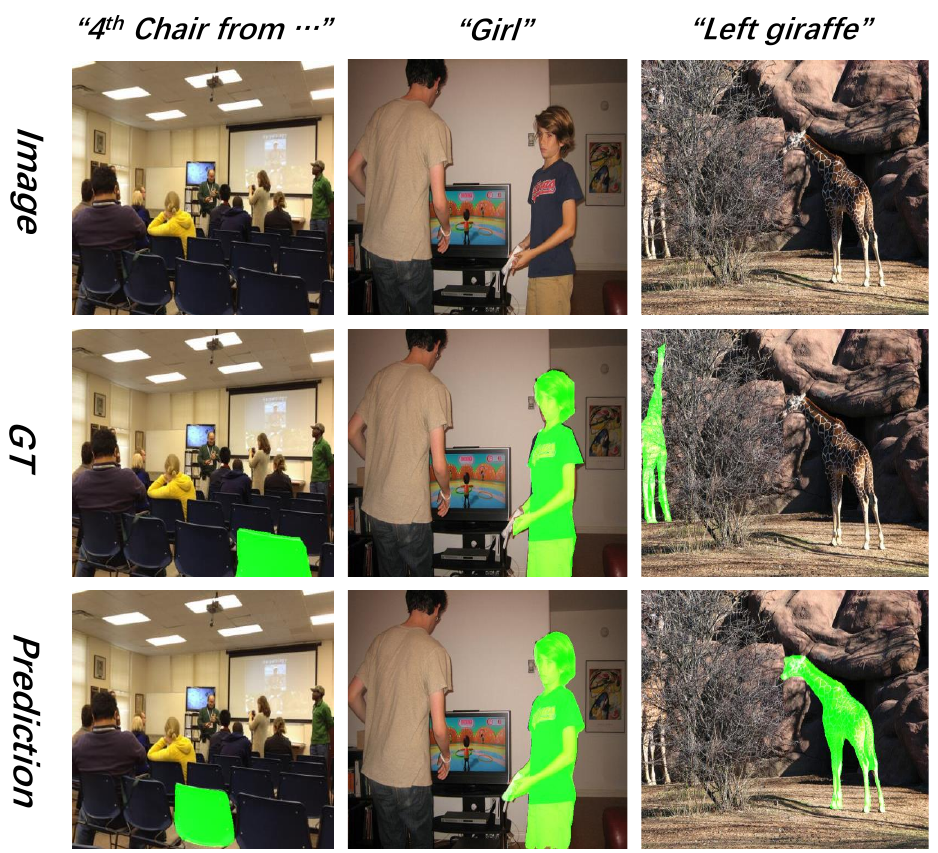}
	\caption{Some failure cases from RefCOCO validation part.}
	\label{failure}
\end{figure}

In Fig. \ref{failure}, some failed results by SwinT-Ours are visualized. In the first and last columns, it can be seen that our network makes mistake in determining the positions of targets. In the future, we can explore to solve the problems by designing new loss that can consider the target existence and resorting to multitask collaboration to integrate referent localization cues, especially in a late fusion way. In the second column, it can be seen that our method also responds to the non-existed target in the image. This limitation is caused by the innate drawbacks of dataset and network logics. In the future, we can propose new datasets and integrate instance existence branch to solve the problem. Moreover, our current network is set to only feed on datasets of RGB and language modalities. Establishing a unified network to handle more binary segmentation subtasks should be able to facilitate the performance of RIS, with dataset imbalance and inductive bias elegantly considered. 

\begin{table}[tbp]  
	\centering
	\caption{The impact of removing decoding stages on the RefCOCO validation set.}
	\resizebox{0.46\textwidth}{!}{
		\setlength\tabcolsep{4pt}
		\renewcommand\arraystretch{1.3}
		\begin{tabular}{l|c|c|c|c|c}
			\toprule
			Method &oIoU$\uparrow$ &mIoU$\uparrow$ &P@0.5$\uparrow$  &P@0.7$\uparrow$  &P@0.9$\uparrow$\\
			\hline
            Full & 72.08 &74.94 &85.39 &75.52 &33.16\\
            -D3 & 70.99 &73.88 &84.07 &74.58 &31.44\\
            -D2\&3 & 70.46 &72.64 &83.68 &72.85 &25.24\\
            -D1\&2\&3& 66.66 &67.11 &79.92 &62.64 &11.33\\
			\bottomrule
	\end{tabular}}	
	\label{tab:stg}
\end{table}

\begin{table}[tbp]  
	\centering
	\caption{Comparison with other fusion methods.}
	\resizebox{0.49\textwidth}{!}{
		\setlength\tabcolsep{4pt}
		\renewcommand\arraystretch{1.3}
		\begin{tabular}{l|c|c|c|c|c}
			\toprule
			Method &oIoU$\uparrow$ &mIoU$\uparrow$ &P@0.5$\uparrow$  &P@0.7$\uparrow$  &P@0.9$\uparrow$\\
			\hline
            ACM~\cite{feng2021encoder} + FPN  & 66.59 &68.47 &78.85 &65.03 &19.08 \\
            PWAM~\cite{yang2022lavt} + FPN & 68.05 &70.70 &81.62 &69.00 &22.16\\
            S\textsuperscript{2}RM + FPN  & 69.57 &71.86 &83.37 &71.49 &23.07\\
            ACM~\cite{feng2021encoder} + CASG & 69.33 &72.39 &83.04 &72.06 &29.19\\
            PWAM~\cite{yang2022lavt} + CASG & 70.87 &73.81 &84.10 &74.09 &30.85\\
            S\textsuperscript{2}RM + CASG & 72.08 &74.94 &85.39 &75.52 &33.16\\
			\bottomrule
	\end{tabular}}	
	\label{tab:cmp_other}
\end{table} 

\section{Conclusion}
In this paper, we propose Spatial Semantic Recurrent Mining (S\textsuperscript{2}RM) to interact vision and language modality features to achieve RIS. The proposed S\textsuperscript{2}RM first generates a distribution-aware yet structure-weal language feature. Then the feature and a pure vision feature are together utilized to co-parse each other in a bidirectional and strucutred way. In different contextual environments, namely vision and language, the S\textsuperscript{2}RM can transport information from near and remote slice layers of one context to the to-be-parsed slice layers from the current context, finally modeling different embeddings relationship globally. The contributions of parsed semantics in S\textsuperscript{2}RM are balanced by the calculated adaptive weights from the balancing step. To reduce computation overhead, the S\textsuperscript{2}RM is installed on top of the feature extractor. We design Cross-scale Abstract Semantic Guided decoder (CASG) to utilize the multiscale semantics in different extracting stages, in which spatial and channel attentions are computed to guide the pure vision feature to supplement effective details and filter out the irrelevant content. Our method performs favorably against other state of the art methods on four challenging datasets.

%-------------------------------------------------------------------------

\bibliographystyle{IEEEtran}
\bibliography{egbib}

\end{document}